\documentclass{article} % For LaTeX2e
\usepackage{amsmath,amsfonts,bm}

\def\eqref#1{equation~\ref{#1}}
\def\1{\bm{1}}

\DeclareMathAlphabet{\mathsfit}{\encodingdefault}{\sfdefault}{m}{sl}
\SetMathAlphabet{\mathsfit}{bold}{\encodingdefault}{\sfdefault}{bx}{n}

\usepackage{graphicx}

\usepackage{placeins} 

\usepackage{hyperref}
\usepackage{url}
\usepackage{booktabs}
\usepackage{tcolorbox}
\usepackage{xcolor}
\usepackage{caption}
\usepackage{geometry}
\usepackage{natbib}

\definecolor{lightgray}{gray}{0.95}
\definecolor{titlepagegray}{gray}{0.94}

\makeatletter
\def\@maketitle{}
\makeatother

\title{}
\author{}
\date{}

\begin{document}

% Custom title page in a box (matching Param-1 style)
\begin{tcolorbox}[
  colback=titlepagegray,
  colframe=titlepagegray,
  boxrule=0pt,
  arc=0pt,
  left=15pt,
  right=15pt,
  top=15pt,
  bottom=15pt,
  width=\textwidth,
  enlarge left by=0mm,
  enlarge right by=0mm
]

% Title
{\sffamily\bfseries\LARGE 
\textsc{AyurParam}: A State-of-the-Art Bilingual Language Model for Ayurveda
}

\vspace{1em}

% Authors
{\bfseries
  Sravan Gorugantu, Mohd Nauman, Vijay Devane, Shyam Pawar, Viraj Thakur, Kundeshwar Pundalik, Piyush Sawarkar, Rohit Saluja, Maunendra Desarkar, Ganesh Ramakrishnan
}

\vspace{0.5em}

% Affiliation
\textit{BharatGen Team}

\vspace{1em}

% Abstract
Current large language models excel at broad, general-purpose tasks, but consistently underperform when exposed to highly specialized domains that require deep cultural, linguistic, and subject-matter expertise. In particular, traditional medical systems such as Ayurveda embody centuries of nuanced textual and clinical knowledge that mainstream LLMs fail to accurately interpret or apply. We introduce \textbf{AyurParam-2.9B}, a domain-specialized, bilingual language model fine-tuned from Param-1-2.9B using an extensive, expertly curated Ayurveda dataset spanning classical texts and clinical guidance. AyurParam's dataset incorporates context-aware, reasoning, and objective-style Q\&A in both English and Hindi, with rigorous annotation protocols for factual precision and instructional clarity. Benchmarked on BhashaBench-Ayur, AyurParam not only surpasses all open-source instruction-tuned models in its size class (1.5--3B parameters), but also demonstrates competitive or superior performance compared to much larger models. The results from AyurParam highlight the necessity for authentic domain adaptation and high-quality supervision in delivering reliable, culturally congruent AI for specialized medical knowledge.

\vspace{1em}

% Date, Correspondence and Logo in a minipage layout
\noindent
\begin{minipage}[b]{0.75\textwidth}
\textbf{Date:} November 5, 2025

\vspace{0.3em}

\textbf{Correspondence:} {kundeshwar.pundalik@tihiitb.org}
\end{minipage}%
\hfill
\begin{minipage}[b]{0.2\textwidth}
\raggedleft
\includegraphics[width=2cm]{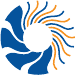} % Adjust path as needed
\end{minipage}

\end{tcolorbox}

\vspace{2em}

\section{Introduction}
Large language models (LLMs) have revolutionized natural language processing by enabling unprecedented understanding, generation, and reasoning across diverse text corpora and tasks. However, despite their general-purpose prowess, these models often fall short when applied to specialized knowledge domains that require deep contextual expertise and cultural awareness. This gap is particularly pronounced in traditional medical systems such as Ayurveda — a comprehensive holistic healthcare system with roots stretching back millennia and rich embedded linguistic, clinical, and philosophical complexities.

Mainstream LLMs are typically trained on extensive but heterogeneous datasets that do not capture the linguistic nuances, semantic specificity, or culturally grounded medical knowledge embedded in Ayurvedic texts and practices. Consequently, unadapted LLMs struggle with accurate interpretation, reasoning, and generation of domain-specific clinical and wellness information, undermining trustworthiness and practical utility in sensitive healthcare contexts. Additionally, many existing generalized models lack bilingual support for Indian languages, further limiting relevance and accessibility for Ayurveda’s practitioner and patient populations.

To bridge these challenges, domain-specialized LLMs have emerged as critical tools—fine-tuned or pretrained on carefully curated, high-quality corpora rich in domain-specific terminology, structured instructional formats, and validated knowledge sources. These tailored models significantly improve accuracy, interpretability, and relevance by internalizing domain conventions, reasoning frameworks, and linguistic subtleties beyond generic LLM capabilities. Applications in clinical decision support, diagnostics, and patient communication have shown promising results, emphasizing the need for linguistic and cultural alignment in medical AI systems.

Building on this paradigm, we present AyurParam, a bilingual large language model specialized for Ayurveda that leverages Param-1-2.9B-Instruct as a base and is fine-tuned on a meticulously assembled dataset comprising digitized classical manuscripts, clinical guidelines, objective assessments, and reasoning-driven queries in both English and Hindi. The dataset is designed with explicit supervision formats, combining dialogue-style prompt-completion pairs and domain expert annotations to enhance instruction-following and factual grounding. AyurParam excels in contextual understanding, reasoning about Ayurvedic principles such as dosha imbalances and samprapti, and delivering culturally nuanced responses across consultation, education, and research use cases.

We comprehensively evaluate AyurParam on the BhashaBench-Ayur benchmark, demonstrating state-of-the-art performance among open-source LLMs in the 1.5–3 billion parameter range and competitive results against substantially larger models. The contributions underscore the transformative potential of domain-specialized LLMs in enabling trustworthy, language-aware AI for traditional medicine, facilitating wider digital accessibility to Ayurveda’s rich heritage while supporting modern clinical workflows.

\section{Related Work}

\textbf{Domain-Specialized Large Language Models}
Early work on foundation models focused on broad-domain generalization, but recent advances reveal the necessity of domain adaptation for deep semantic understanding and accurate task execution in specialized fields [Ling et al., 2023]. The comprehensive survey by Ling et al. (2023) categorizes domain adaptation strategies into direct fine-tuning, targeted pre-training, and external data augmentation, each facilitating models with richer domain vocabulary, reasoning methods, and error resilience. Biomedical LLMs such as PubMedGPT and BioGPT employ massive medical corpora to internalize discipline-specific knowledge and terminology, yielding superior performance in medical QA and summarization tasks compared to generalized LLMs. Fine-tuning on well-structured domain datasets and leveraging synthetic instruction instances are critical for achieving both higher accuracy and trustworthiness in high-stakes applications.

\textbf{Instruction Tuning and Task Adaptation}
Instruction tuning has emerged as a pivotal method to align LLMs with human expectations, safety constraints, and domain ontologies [Alaa et al., 2025; IBM, 2024]. Recent studies explore multi-task and prompt-style frameworks, including FLAN, SuperNI, and Self-Instruct, to expose models to diverse instructional templates and domain-conditional objectives [Wei et al., 2021; Wang et al., 2022]. Meta-tuning, domain-aware task conditioning, and reinforcement learning (e.g., RLHF, DPO) offer further enhancements for flexibility, factual consistency, and safer outputs. In educational and scientific fields, instruction-tuned LLMs exhibit dramatically improved reasoning and answer quality, but achieving similar success in medical and traditional knowledge domains demands curated dataset design, multilingual coverage, and expert validation.

\textbf{Medical and Multilingual LLM Benchmarks}
Medical and healthcare adaptation of LLMs is rapidly advancing, especially where multilingual coverage and cultural grounding are needed [Qiu et al., 2024]. Benchmarks such as MMedBench and Swedish Medical LLM Benchmark have been introduced to systematically evaluate clinical reliability, multilingual QA performance, and safety in realistic environments. Adoption of domain-specific benchmarks and layered evaluation protocols is advocated to address limitations in existing medical leaderboards and guide ethical, scalable AI deployment in sensitive contexts [Alaa et al., 2025]. Studies consistently reveal that models fine-tuned with localized, expert-curated corpora outperform generalist LLMs and demonstrate practical utility in clinical consulting, wellness, and patient communication.

\textbf{AI for Ayurveda and Traditional Medicine}
While most attention has centered on Western biomedical LLMs, a small but growing body of work addresses the challenges of adapting AI and LLMs to traditional medical systems like Ayurveda [Padia et al., 2025; Rathor et al., 2024]. Early rule-based and narrow ML applications offered limited interpretability and reasoning, but newer models—including AyurGPT and IRGPT—leverage domain-specific pre-training and multilingual instruction-tuning to support consultation and diagnosis across languages. These models process Sanskrit and regional linguistic data, enhance reasoning about dosha, dhatu, and treatment logic, and are benchmarked with domain-specific metrics, yet their scope and quality remain underexplored at scale. The work done builds on these foundations, presenting AyurParam as the first bilingual, instruction-tuned LLM extensively benchmarked for authentic, context-rich performance in Ayurveda.

%% Please note that we have introduced automatic line number generation
%% into the style file for \LaTeXe. This is to help reviewers
%% refer to specific lines of the paper when they make their comments. Please do
%% NOT refer to these line numbers in your paper as they will be removed from the
%% style file for the final version of accepted papers.

% \section{Data} TBD by Nauman

\section{Data Preparation}
\label{sec:data-prep}

\begin{figure}[h]
    \centering
    \includegraphics[width=0.75\linewidth]{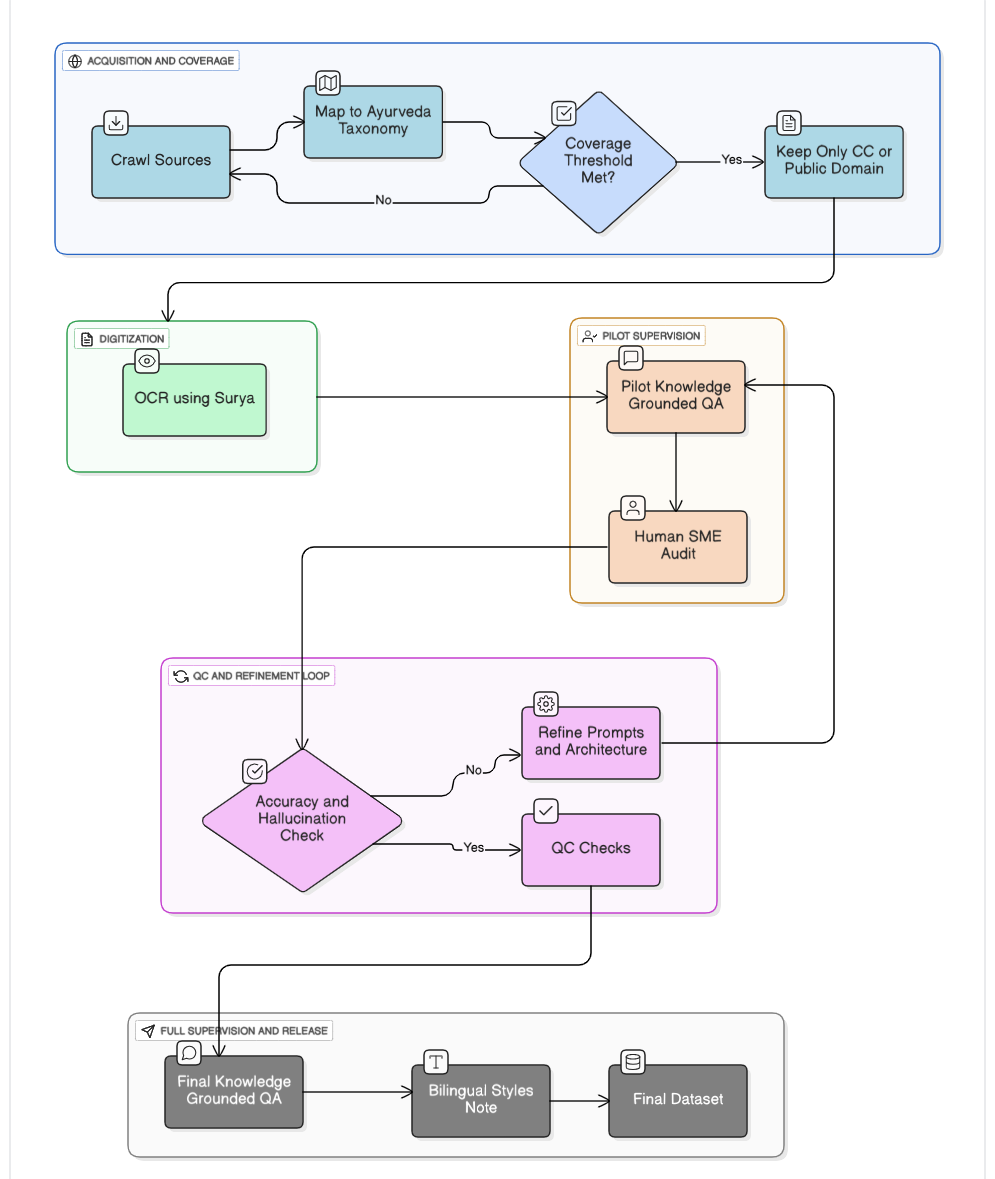}
    \caption{Data preparation pipeline}
    \label{fig:data_prep_pipeline}
\end{figure}

Our data preparation methodology follows a systematic pipeline encompassing taxonomy establishment, corpus collection, OCR processing, quality assurance, and knowledge-grounded Q\&A generation (Figure~\ref{fig:data_prep_pipeline}) . Each stage incorporates domain-specific constraints and validation protocols to ensure the resulting dataset meets the rigorous requirements for specialized Ayurvedic instruction tuning.

\FloatBarrier
\FloatBarrier

\subsection{Taxonomy of Ayurvedic Domains}
Before collecting or processing texts, we established a curriculum–aligned taxonomy to ensure that
all major branches of Ayurveda are represented. This step prevents over-representation of easily
available material (e.g., Panchakarma manuals) and ensures inclusion of essential domains
that are often underrepresented in digital form. 

The taxonomy was derived from the official \textit{BAMS} undergraduate curriculum, postgraduate
\textit{MD/MS} specializations, and canonical compendia (Charaka, Sushruta, Ashtanga Hridaya,
Kashyapa Samhita). It serves three purposes:
\begin{enumerate}
  \item Acts as a retrieval lens when querying archives in Devanagari, IAST, and English transliteration.  
  \item Enforces per-domain quotas to maintain balance and prevent skew.  
  \item Defines strata for downstream evaluation and error analysis.  
\end{enumerate}

\noindent\textbf{Domains included:}
\begin{itemize}
  \item \textit{Foundations}: Ashtang Hridaya, Padarth Vigyan, selected Sanskrit commentaries.  
  \item \textit{Anatomy and Physiology}: Rachana Shaarir, Kriya Shaarir.  
  \item \textit{Classical Compendia}: Charaka Samhita, Sushruta Samhita.  
  \item \textit{Clinical Disciplines}: Kayachikitsa, Panchakarma, Shalya Tantra, Shalakya Tantra.  
  \item \textit{Pharmacology and Formulations}: Dravyaguna, Rasa Shastra, Bhaishajya Kalpana.  
  \item \textit{Pathology and Toxicology}: Rog Nidan, Agad Tantra.  
  \item \textit{Specialties}: Kaumarbhritya (Balrog), Strirog and Prasuti Tantra, Swasthavritta.  
\end{itemize}

This taxonomy acted as the scaffold against which all subsequent collection, license filtering,
OCR, and Q\&A generation were aligned.

\subsection{Corpus Collection and License Governance}
\label{sec:acquisition-license}

\textbf{Acquisition.} We collected approximately 1,000 Books and doc spanning classical compendia, modern
commentaries, and clinical manuals. Of these, $\sim$600 were sourced from public digital archives and
open libraries (e.g., Archive.org, eGangotri), while $\sim$400 were drawn from institutional or
government repositories. The collection spans Sanskrit originals, Hindi and Marathi translations,
and English editions, ensuring bilingual coverage while preserving classical terminology.
In total, the corpus comprises $\sim$150{,}000 pages ($\sim$54.5M words), forming one of the largest curated Ayurvedic text datasets.

In total, the corpus comprises $\sim$150{,}000 pages ($\sim$54.5M words), forming one of the largest curated Ayurvedic text datasets. 

\begin{figure}[h]
    \centering
    \includegraphics[width=0.6\linewidth]{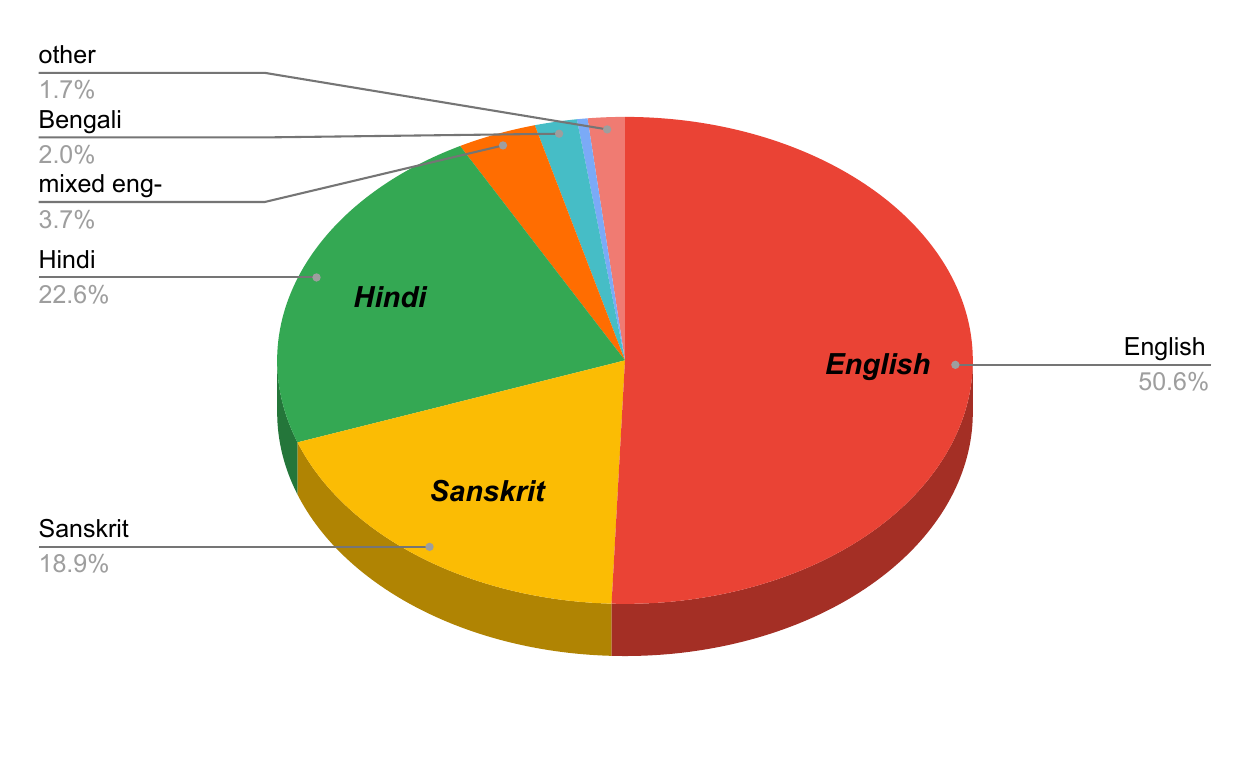}  % Your pie chart
    \caption{Language-wise distribution of the collected corpus across Sanskrit, Hindi, Marathi, and English sources}
    \label{fig:language_distribution}
\end{figure}

Figure~\ref{fig:language_distribution} shows the language-wise distribution of crawled source documents, with PDFs collected in Sanskrit (Devanagari script), Hindi, Marathi, and English, ensuring comprehensive multilingual coverage of Ayurvedic literature.

\FloatBarrier
\FloatBarrier

\textbf{License governance.} Each item was catalogued into a \emph{license ledger} that records edition-level
metadata, including title, authors, translators, publication year, source URL, language tags, and
declared rights. Only texts with clear open licenses (e.g., CC0, CC-BY, public domain) were retained
for training. Works with ambiguous or restrictive licenses were preserved as \emph{shadow entries}—
metadata retained for completeness but text excluded from supervision. This approach preserves
reproducibility while respecting legal and ethical constraints.

\textbf{Duplicate handling.} Multiple editions of the same compendium are frequent in Ayurveda
(e.g., Charaka Samhita with different commentaries). To avoid redundancy, we applied page-level
near-duplicate detection using character n-gram hashing and MinHash signatures. Unique editions were
retained, and their relationships were cross-linked in metadata to support lineage tracking.

This two-step process of broad collection followed by strict license filtering ensured that the final
dataset is both comprehensive and reproducible, while remaining legally and ethically compliant.

\subsection{OCR and Post-OCR Processing}
\label{sec:ocr}

\textbf{Pre-processing.} Many collected works were available only as scanned PDFs or images with
significant variance in quality. Before OCR, each page was conditioned through deskewing,
contrast normalization, and margin cropping. This ensured more consistent recognition across
heterogeneous sources.

\textbf{OCR engine.} For Indic scripts (Devanagari: Sanskrit, Hindi, Marathi), we employed
\textbf{Surya OCR}, which provides state-of-the-art performance on multilingual text lines.
English pages were processed in the same pipeline for consistency. Page-level OCR confidence
scores were recorded as metadata for subsequent filtering.

\textbf{Normalization.} Post-OCR text was standardized through several layers of cleaning:
\begin{itemize}
  \item Unicode NFC normalization, repair of common Devanagari ligatures and numerals.  
  \item Removal of headers, footers, and marginalia to reduce noise.  
  \item De-hyphenation and whitespace normalization.  
  \item Segmentation of passages with language tags (\texttt{san-Deva}, \texttt{hi-Deva}, \texttt{en-Latn}).  
  \item Alignment with canonical divisions when detectable (e.g., Sūtrasthāna, Nidānasthāna,
        Vimānasthāna, Uttaratantra).  
\end{itemize}

\textbf{Quality assurance.} We monitored OCR accuracy using a combination of automatic and heuristic
checks: (i) mean and median confidence per page; (ii) crude character error rate estimation via
lexicon-free sampling; (iii) Indic-specific heuristics such as akshara merges or danda detection.
Low-confidence pages were flagged for exclusion or routed to stricter cleaning.

This stage yielded a normalized, linguistically tagged corpus suitable for downstream Q\&A generation,
with provenance and quality metadata attached at the page level.

\subsection{Knowledge-Grounded Data Generation}
\label{sec:qna}

\textbf{Knowledge-grounded synthesis.} Cleaned passages were transformed into supervised training
examples using high-capacity LLMs (primarily \textbf{Qwen-3 235B}) under strict constraints:
responses were required to be derivable from the provided span, with no external elaboration or
prescriptive advice.

\textbf{Human-in-the-loop calibration.} To improve reliability, ten representative books were
manually reviewed by Ayurveda practitioners. For each book, 50--300 pages were randomly sampled and
the corresponding Q\&A pairs analyzed. Experts identified over-generalization, implicit assumptions,
and unsupported reasoning, which informed iterative refinements to the synthesis policy. This loop
significantly reduced hallucination and improved epistemic fidelity.

\textbf{Quality assurance.} Reliability was enforced through a staged validation pipeline:  
\begin{itemize}
  \item \emph{Rule-based filters:} Every item was checked for JSON schema validity, minimum/maximum
  answer length, banned phrases (e.g., prescriptive treatment advice), and symbol consistency.  
  \item \emph{Evidence anchoring:} Answers were required to cite support spans from the passage. We
  measured lexical overlap and coverage ratios to detect unsupported generations.  
  \item \emph{Selective LLM adjudication:} Only uncertain cases—those failing thresholds or flagged
  by overlap heuristics—were escalated to an LLM-as-judge for groundedness and contradiction checks.  
  \item \emph{Targeted human audits:} Stratified samples from uncertain cases, low-OCR-confidence
  pages, and high-stakes domains (e.g., \emph{Śalya}, \emph{Śālākya}) were reviewed by practitioners,
  with inter-annotator agreement monitored for consistency.  
\end{itemize}

\textbf{Final dataset.} The resulting corpus contained approximately 4.75M grounded Q\&A pairs,
distributed as:
\begin{itemize}
  \item \textbf{Q\&A pair (EN + HI)}: $\sim$1.27M pairs.  
  \item \textbf{Objective/MCQ}: $\sim$0.9M pairs.  
  \item \textbf{Multi-turn reasoning}: $\sim$1.51M pairs.  
  \item \textbf{Contextual Q\&A (span-level comprehension)}: $\sim$1.07M pairs.  
\end{itemize}

\begin{figure}[h]
    \centering
    \includegraphics[width=0.55\linewidth]{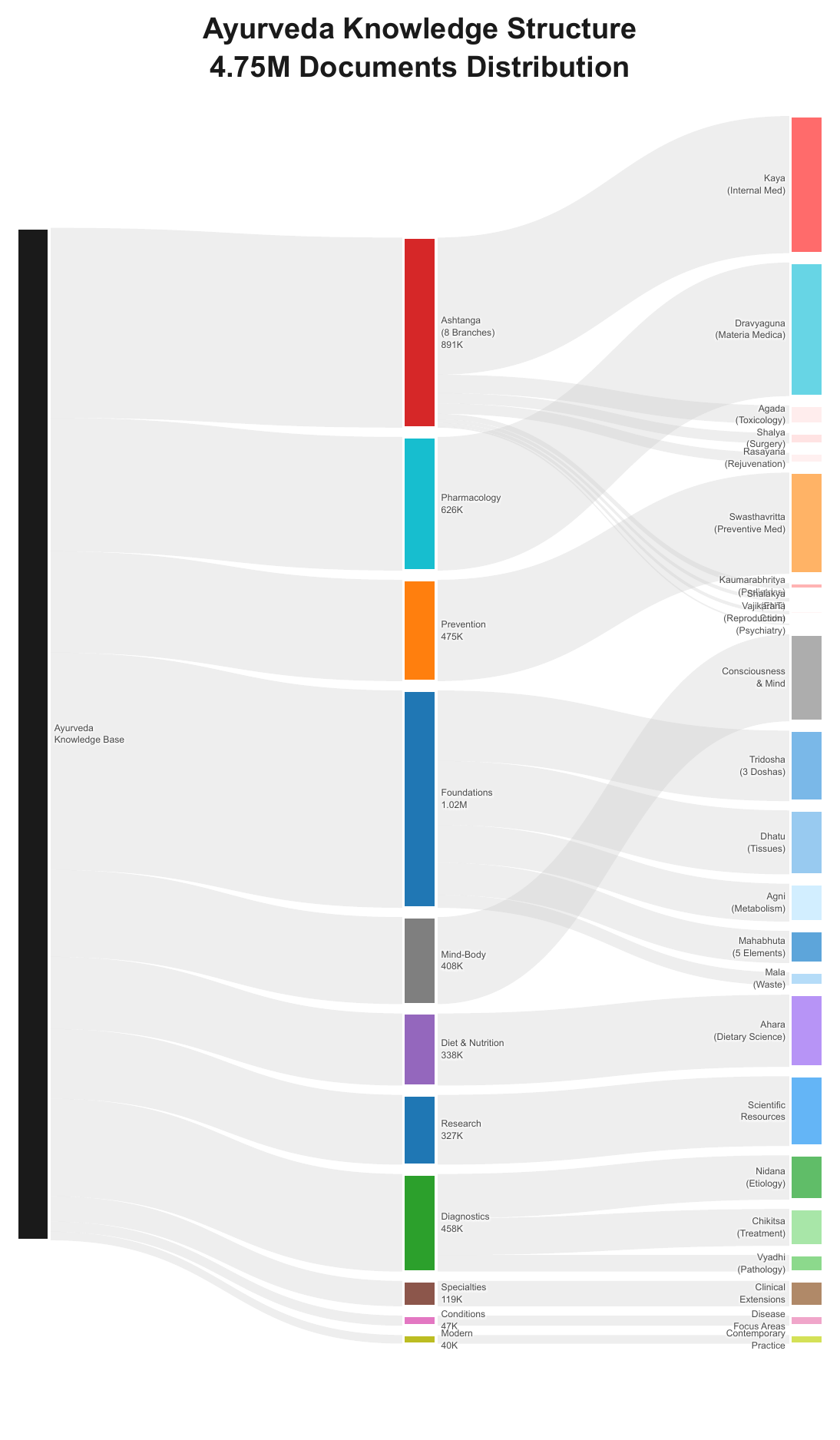}
    \caption{QA-Distribution}
    \label{fig:qa-distribution}
\end{figure}

This balanced coverage ensures both breadth (terminology, principles, clinical applications) and
depth (reasoning, multi-turn consistency). As shown in Figure~\ref{fig:qa-distribution},
the dataset exhibits a balanced distribution across question types.

\FloatBarrier
\FloatBarrier

\subsection{Ethical and Legal Considerations}
\label{sec:ethics}

All primary texts were sourced from public digital archives (e.g., Archive.org, eGangotri) and 
institutional repositories that explicitly provide open access or carry permissive licenses. 
A license filter was applied to exclude restricted or unclear materials, ensuring that the final 
corpus is legally redistributable for research and educational purposes. 

Sensitive sources such as personal health records, unpublished manuscripts, and private clinical 
notes were explicitly excluded to prevent leakage of private or identifiable information. 
Additionally, Q\&A generation was framed to support research and education rather than clinical 
prescription, reflecting the model’s intended use. 

We note that classical Ayurvedic texts themselves embed cultural and historical biases 
(e.g., gender roles in physician narratives). While AyurParam may reflect such tendencies, 
the instruction-tuning framework reduces prescriptive phrasing and emphasizes factual recall.

\section{Training}

\subsection{Training Data}

The supervised fine-tuning (SFT) dataset for AyurParam was built from the curated Ayurveda corpus discussed in Section 3. The focus during dataset construction was twofold: ensuring broad coverage across classical Ayurvedic texts, clinical subdomains, and wellness knowledge, and enabling strong instruction-following capabilities.
\paragraph{Conversational Alignment Data.} 
For conversational alignment, the data was organized in a dialogue-style format using 
\texttt{<user>} and \texttt{<assistant>} markers. Each \texttt{<user>} tag contained the query 
(question, prompt, or instruction), while the \texttt{<assistant>} tag held the model’s reply. 
To provide explicit supervision, responses were further enclosed within 
\texttt{<actual\_response>} and \texttt{</actual\_response>}, clearly indicating the ground-truth 
output for fine-tuning. This structure helped maintain a clean separation between prompts, 
user inputs, and reference answers.  

The dataset integrated multiple generation methods, including:  
\begin{itemize}
    \item \textbf{Context-based Q\&A:} Questions tied to page-level or passage-level content 
    from traditional Ayurvedic manuscripts.
    \item \textbf{Instructional Prompts:} Generic tasks designed to elicit structured reasoning.
    \item \textbf{Situational Dialogue:} Conversational, scenario-driven exchanges to improve alignment.

   \item \textbf{Objective-style tasks:} Multiple-choice, fill-in-the-blank, and structured assessments for clinical and academic use.

    \item \textbf{Reasoning-driven questions:} Prompts requiring analytical thinking around Ayurvedic principles such as dosha imbalances, samprapti (disease progression), and treatment logic etc.,

    \item \textbf{General knowledge Q\&A:} Covering terminology, principles, and practices in both English and Hindi.
    
\end{itemize}

Altogether, the training set contained about ~4.75 million examples, spanning single-turn and multi-turn dialogues. This design not only supported factual recall but also strengthened reasoning and inference skills—key requirements for a specialized conversational agent in Ayurveda.

\subsection{Training Setup}

We fine-tuned the \texttt{Param-1-2.9B} \cite{param1} base model using the Hugging Face TRL framework in a supervised fine-tuning (SFT) configuration. Training was performed on a multi-node NVIDIA H100 GPU cluster with \texttt{torchrun}. The main hyperparameters and settings are summarized below:

\begin{itemize}
    \item \textbf{Global batch size:} 1024 (micro-batch size = 4; gradient accumulation = 32)
    \item \textbf{Learning rate:} $5 \times 10^{-6}$ with linear decay and warmup
    \item \textbf{Training epochs:} 3
    \item \textbf{Precision:} bfloat16 mixed precision
    \item \textbf{Vocabulary:} 256k tokens + 6 task-specific tokens (\texttt{<user>}, \texttt{<assistant>}, \texttt{<context>}, \texttt{<system\_prompt>}, \texttt{<actual\_response>}, \texttt{</actual\_response>})
\end{itemize}

The supervised fine-tuning corpus consisted of approximately 4.75M samples. Training required roughly two days on a single H100 node. To better support both single-turn and multi-turn Ayurvedic instruction-following, we employed custom bilingual templates (English and Hindi).

\section{Model Performance}

To rigorously evaluate domain specialization, we benchmarked \texttt{AyurParam-2.9B} on \texttt{BhashaBench-Ayur} (BBA), India's first large-scale evaluation suite for Ayurvedic AI systems \cite{bhashabench2025}. BBA consists of 14,963 exam-style questions spanning 15+ subject domains, covering both English (9,348 questions) and Hindi (5,615 questions). The dataset integrates authentic government and institutional examinations across India, reflecting the same rigor and breadth expected of BAMS graduates and postgraduate Ayurvedic training. It evaluates factual recall, clinical reasoning, therapeutic principles, and interpretability through diverse question formats (MCQ, assertion-reasoning, fill-in-the-blanks, and match-the-column).

% Table 1: Overall Performance
\begin{table}[htbp]
\centering
\caption{Overall performance comparison on BBA dataset. Results show accuracy (\%) across different model sizes. Our AyurParam-2.9B-Instruct model achieves the best performance among similar-sized models and competitive results compared to much larger models.}
\label{tab:overall_performance}
\begin{tabular}{@{}lccc@{}}
\toprule
\multicolumn{4}{c}{\textbf{Similar Range Models}} \\
\midrule
\textbf{Model} & \textbf{BBA} & \textbf{BBA English} & \textbf{BBA Hindi} \\
\midrule
\textbf{AyurParam-2.9B-Instruct} & \textbf{39.97} & \textbf{41.12} & \textbf{38.04} \\
Llama-3.2-3B-Instruct & 33.20 & 35.31 & 29.67 \\
Qwen2.5-3B-Instruct & 32.68 & 35.22 & 28.46 \\
granite-3.1-2B & 31.10 & 33.39 & 27.30 \\
gemma-2-2B-it & 28.40 & 29.38 & 26.79 \\
Llama-3.2-1B-Instruct & 26.41 & 26.77 & 25.82 \\
\midrule
\multicolumn{4}{c}{\textbf{Larger Models}} \\
\midrule
\textbf{AyurParam-2.9B-Instruct} & \textbf{39.97} & \textbf{41.12} & \textbf{38.04} \\
gemma-2-27B-it & 37.99 & 40.45 & 33.89 \\
Pangea-7B & 37.41 & 40.69 & 31.93 \\
gpt-oss-20B & 36.34 & 38.30 & 33.09 \\
Indic-gemma-7B-Navarasa-2.0 & 35.13 & 37.12 & 31.83 \\
Llama-3.1-8B-Instruct & 34.76 & 36.86 & 31.26 \\
Nemotron-4-Mini-Hindi-4B-Instruct & 33.54 & 33.38 & 33.82 \\
aya-23-8B & 31.97 & 33.84 & 28.87 \\
\bottomrule
\end{tabular}
\end{table}

On this benchmark, \texttt{AyurParam-2.9B} achieved state-of-the-art accuracy among models in the 1.5--3B parameter range, outperforming instruction-tuned baselines such as LLaMA-3.2-3B and Qwen2.5-3B. Notably, despite being significantly smaller, \texttt{AyurParam-2.9B} demonstrated competitive performance with models in the 7B--27B range, underscoring the efficiency of targeted domain specialization (Table~\ref{tab:overall_performance}).

% Table 2: Question Difficulty Analysis
\begin{table}[htbp]
\centering
\caption{Performance breakdown by question difficulty. Results demonstrate that AyurParam-2.9B-Instruct maintains strong performance across all difficulty levels, with particularly notable results on easy questions.}
\label{tab:difficulty_analysis}
\begin{tabular}{@{}lcccccc@{}}
\toprule
\multicolumn{7}{c}{\textbf{Similar Range Models}} \\
\midrule
\textbf{Difficulty} & \textbf{AyurParam-2.9B} & \textbf{Llama-3B} & \textbf{Qwen-3B} & \textbf{Granite-2B} & \textbf{Gemma-2B} & \textbf{Llama-1B} \\
\midrule
Easy & \textbf{43.93} & 36.42 & 35.55 & 33.90 & 29.96 & 27.44 \\
Medium & \textbf{35.95} & 29.66 & 29.57 & 28.06 & 26.83 & 25.23 \\
Hard & \textbf{31.21} & 28.51 & 28.23 & 26.81 & 24.96 & 25.39 \\
\bottomrule
\end{tabular}

\vspace{0.5cm}

\begin{tabular}{@{}lcccccc@{}}
\toprule
\multicolumn{7}{c}{\textbf{Larger Models}} \\
\midrule
\textbf{Difficulty} & \textbf{AyurParam-2.9B} & \textbf{Gemma-27B} & \textbf{Pangea-7B} & \textbf{Llama-8B} & \textbf{Indic-7B} & \textbf{Aya-8B} \\
\midrule
Easy & \textbf{43.93} & 43.47 & 41.45 & 39.43 & 38.54 & 35.51 \\
Medium & \textbf{35.95} & 31.90 & 32.94 & 29.36 & 31.72 & 28.29 \\
Hard & 31.21 & 30.78 & \textbf{31.77} & 30.50 & 27.23 & 25.11 \\
\bottomrule
\end{tabular}
\end{table}

The model showed strong gains on clinically relevant domains such as \emph{Kayachikitsa} and \emph{Dravyaguna}, while also maintaining robust performance in medium- and high-difficulty questions, where general-purpose models often struggled (Table~\ref{tab:difficulty_analysis}). These results highlight that high-quality, domain-specific supervision and carefully designed instruction-tuning protocols enable small- and mid-scale models to achieve outsized gains in specialized benchmarks.

Evaluation on BBA also revealed limitations: performance in Hindi lagged behind English, and reasoning-intensive domains such as \emph{Panchakarma \& Rasayana} and \emph{Ayurvedic Literature} remained challenging. This points to the need for improved multilingual coverage and deeper alignment with classical knowledge representations, which we consider essential for the next phase of research.

% Table 3: Question Type Analysis
\begin{table}[htbp]
\centering
\caption{Performance analysis across different question types. AyurParam-2.9B-Instruct shows strong performance on MCQ questions and competitive results across all question formats.}
\label{tab:question_type}
\begin{tabular}{@{}lccccc@{}}
\toprule
\multicolumn{6}{c}{\textbf{Similar Range Models}} \\
\midrule
\textbf{Type} & \textbf{Llama-1B} & \textbf{Qwen-3B} & \textbf{Llama-3B} & \textbf{AyurParam-2.9B} & \textbf{Gemma-2B} \\
\midrule
Assert./Reason. & 59.26 & 51.85 & 40.74 & 44.44 & 33.33 \\
Fill blanks & 26.97 & 29.21 & 34.83 & 29.78 & 32.02 \\
MCQ & 26.34 & 32.70 & 33.17 & 40.12 & 28.33 \\
Match col. & 26.83 & 29.27 & 29.27 & 24.39 & 36.59 \\
\bottomrule
\end{tabular}

\vspace{0.5cm}

\begin{tabular}{@{}lccccc@{}}
\toprule
\multicolumn{6}{c}{\textbf{Larger Models}} \\
\midrule
\textbf{Type} & \textbf{Pangea-7B} & \textbf{Gemma-27B} & \textbf{AyurParam-2.9B} & \textbf{Llama-8B} & \textbf{Indic-7B} \\
\midrule
Assert./Reason. & 62.96 & 55.56 & 44.44 & 29.63 & 59.26 \\
Fill blanks & 24.16 & 35.96 & 29.78 & 26.97 & 35.39 \\
MCQ & 37.53 & 37.98 & 40.12 & 34.83 & 35.10 \\
Match col. & 34.15 & 39.02 & 24.39 & 46.34 & 31.71 \\
\bottomrule
\end{tabular}
\end{table}

The question type analysis reveals nuanced performance patterns across different evaluation formats (Table~\ref{tab:question_type}). \texttt{AyurParam-2.9B} excels particularly in multiple-choice questions (MCQ), achieving the highest accuracy (40.12\%) among all compared models, including those with significantly more parameters. This strength in MCQ performance is especially valuable for Ayurvedic assessment, as it reflects the model's ability to discriminate between closely related concepts and therapeutic approaches—a critical skill for clinical decision-making.

\section{Conclusion}

This work introduced \textbf{AyurParam}, a bilingual, domain-specialized large language model for Ayurveda, fine-tuned from the Param-1-2.9B-Instruct foundation model. By leveraging a rigorously curated corpus of classical texts, clinical manuals, and bilingual instructional data, AyurParam achieves high accuracy across a diverse set of question types and difficulty levels, outperforming similar-sized open-source models and remaining competitive with much larger models. \\ \\ Extensive evaluation on \textbf{BhashaBench-Ayur} demonstrates that careful domain adaptation, high-quality supervision, and culturally grounded pretraining are critical for enabling small- to mid-scale models to perform reliably on complex, specialized knowledge tasks. In particular, AyurParam shows robust performance in multiple-choice questions, reasoning-intensive prompts, and multi-turn Q\&A, highlighting its potential as a practical tool for Ayurvedic education, research, and clinical knowledge support. \\ \\ The study underscores the value of domain-specialized, multilingual LLMs in bridging gaps between traditional knowledge systems and modern AI. Future work will focus on enhancing reasoning in high-difficulty scenarios, improving Hindi and Sanskrit understanding, and expanding coverage to underrepresented Ayurvedic subdomains, further advancing trustworthy, culturally aligned AI for traditional medicine.

\section{Limitations}

Despite strong performance on the BhashaBench-Ayur benchmark, AyurParam has several limitations that warrant acknowledgment:

\paragraph{Temporal Coverage and Contemporary Knowledge.}
The training corpus consists primarily of classical Ayurvedic texts and open-source educational materials digitized before 2024. As a result, AyurParam has limited exposure to recent research developments, contemporary clinical practices, and ongoing advances in Ayurveda. This temporal gap is common in domain-specialized models~\citep{ling2023,alaa2025} but particularly consequential in medical domains where evidence and practices evolve continuously.

\paragraph{Language Performance Gap.}
While AyurParam supports both English and Hindi, performance on Hindi queries lags behind English (38.04\% vs. 41.12\% accuracy, Table~\ref{tab:overall_performance}). This disparity suggests insufficient representation of Hindi content in the training corpus and highlights the ongoing challenge of building truly balanced multilingual models for specialized domains~\citep{qiu2024}.

\paragraph{Evaluation Scope and Methodology.}
Our evaluation relies exclusively on structured exam-style questions from BhashaBench-Ayur. While this benchmark provides rigorous assessment of factual knowledge and reasoning, it does not capture other critical dimensions such as: (i) open-ended generation quality; (ii) clinical reasoning in realistic consultation scenarios; (iii) safety and appropriateness of generated advice; and (iv) practitioner acceptance and usability. Human evaluation by Ayurvedic experts would provide complementary insights but was beyond the scope of this work~\citep{alaa2025}.

\paragraph{Safety and Ethical Guardrails.}
Although trained on factual, expert-curated content, AyurParam lacks explicit safety mechanisms to prevent generation of inappropriate, unsafe, or potentially harmful medical advice. Prior work in medical LLMs emphasizes the necessity of robust guardrails and clinical validation before deployment~\citep{qiu2024,alaa2025}. Current responses are knowledge-grounded but not clinically validated.

\paragraph{Personalization and Context-Awareness.}
AyurParam does not account for individual patient histories, contraindications, or personalized health contexts in its responses. While this aligns with its design as an educational and reference tool, practical clinical utility would benefit from patient-specific reasoning capabilities.

\paragraph{Data Licensing and Diversity.}
The corpus is limited to open-access texts from public repositories such as Archive.org, eGangotri, and NDLI~\citep{archiveorg,egangotri,ndli}. Incorporating licensed clinical databases, contemporary peer-reviewed literature, and modern practice guidelines would enhance both coverage and reliability but requires navigating complex licensing and copyright considerations.

\section{Future Work}

Several directions emerge naturally from these limitations:

\paragraph{Incorporating Contemporary Knowledge.}
Future iterations should integrate recent research publications, institutional clinical guidelines, and emerging practices to maintain temporal relevance. This may involve continual learning frameworks or periodic retraining cycles similar to approaches used in general medical LLMs~\citep{qiu2024}.

\paragraph{Improving Multilingual Balance.}
Addressing the Hindi performance gap requires targeted data augmentation, improved tokenization strategies, and possibly language-specific fine-tuning. Techniques from multilingual NLP research~\citep{ai4bharat_corpora,indicnlp,mc4} could be adapted to the Ayurvedic domain.

\paragraph{Human Evaluation and Clinical Validation.}
Systematic evaluation by Ayurvedic practitioners is essential to assess clinical utility, safety, and appropriateness of generated responses. This could follow frameworks established for medical AI systems~\citep{alaa2025} and include qualitative analysis of open-ended consultations.

\paragraph{Safety Mechanisms and Guardrails.}
Future versions should implement explicit safety layers to detect and prevent generation of harmful advice, incorporating lessons from instruction-tuning and alignment research~\citep{instructgpt,wang2022,alpaca}. This includes disclaimer generation, uncertainty quantification, and refusal mechanisms for out-of-scope queries.

\paragraph{Enhanced Data Sources.}
Expanding the corpus to include licensed clinical data, modern formulations, and validated treatment protocols would improve both accuracy and practical utility. Collaboration with Ayurvedic institutions and regulatory bodies could facilitate access to high-quality, authoritative sources.

\paragraph{Accessibility and User Adaptation.}
Developing mechanisms to adapt response complexity based on user expertise (practitioners vs. patients vs. students) would enhance practical utility across diverse use cases while maintaining scientific accuracy. \\ \\ By addressing these limitations, future research can advance AyurParam toward a more comprehensive, safe, and clinically validated tool for Ayurvedic knowledge dissemination and education.

\bibliographystyle{plain}

\begin{thebibliography}{99}

% --- BharatGenAI Models & Benchmarks ---
\bibitem{bharatgen2025ayurparam}
BharatGenAI. (2025). \emph{AyurParam: Domain-specialized Ayurvedic LLM}. Hugging Face. 
\url{https://huggingface.co/bharatgenai/AyurParam}

\bibitem{bhashabench2025}
BharatGen Research Team. (2025). \emph{BhashaBench-Ayur (BBA): Pioneering India’s Ayurvedic AI Benchmark}. Hugging Face Datasets. \url{https://huggingface.co/datasets/bharatgenai/bhashabench-ayur}

\bibitem{param1}
 BharatGen Research Team. (2025). \emph{PARAM-1 BharatGen 2.9B Model}. arXiv preprint arXiv:2507.13390. \url{https://arxiv.org/abs/2507.13390}


% --- Domain Specialization & Medical LLMs ---
\bibitem{ling2023}
Ling, C., et al. (2023). Domain Specialization as the Key to Make Large Language Models Efficient, Reliable, and Explainable. 
\emph{arXiv preprint arXiv:2305.18703}.

\bibitem{alaa2025}
Alaa, A., et al. (2025). Rethinking Medical Benchmarks for Large Language Models. 
\emph{arXiv preprint arXiv:2508.04325}.

\bibitem{qiu2024}
Qiu, P., et al. (2024). Towards Building Multilingual Language Model for Medicine. 
\emph{Nature Communications}. \url{https://doi.org/10.1038/s41467-024-52417-z}

% --- Instruction Tuning & Alignment ---
\bibitem{ibm2024}
IBM Research. (2024). InstructLab: Large-scale Alignment for chatBots (LAB). 
\emph{Red Hat Blog}. \url{https://www.redhat.com/en/topics/ai/what-is-instructlab}

\bibitem{wei2021}
Wei, J., et al. (2021). Finetuned Language Models Are Zero-Shot Learners. 
\emph{arXiv preprint arXiv:2109.01652}.

\bibitem{wang2022}
Wang, Y., et al. (2022). Self-Instruct: Aligning Language Models with Self-Generated Instructions. 
\emph{arXiv preprint arXiv:2212.10560}. \url{https://arxiv.org/abs/2212.10560} 

\bibitem{instructgpt}
Ouyang, L., et al. (2022). Training Language Models to Follow Instructions with Human Feedback (InstructGPT). 
\emph{NeurIPS}.

\bibitem{alpaca}
Taori, R., et al. (2023). Stanford Alpaca: An Instruction-Following LLaMA Model. 
\emph{Stanford CRFM Blog}.

\bibitem{dolly}
Databricks. (2023). Databricks Dolly: Democratizing the Magic of ChatGPT. 
\emph{Databricks Blog}.

\bibitem{sharegpt}
ShareGPT contributors. (2023). ShareGPT Dataset. \url{https://sharegpt.com}

\bibitem{ultrachat}
Ding, N., et al. (2023). UltraChat: A Large-Scale Chat Dataset for Instruction Tuning. 
\emph{arXiv preprint arXiv:2305.14233}.

% --- Ayurveda & AI ---
\bibitem{padia2025}
Padia, A., et al. (2025). AyurGPT: Fine-Tuning a Medical LLM for Multilingual Ayurveda Consultations. 
\emph{NLP Summit Presentation}. 
\url{https://www.nlpsummit.org/ayur-gtp-fine-tuning-a-medical-llm-for-multilingual-ayurveda-consultations/}

\bibitem{rathor2024}
Rathor, C., et al. (2024). Artificial Intelligence in Ayurveda: A Simple Overview. 
\emph{Journal of Ayurveda and Integrative Medical Sciences}. 
\url{https://www.jaims.in/jaims/article/view/4040}

% --- Classical Ayurveda Texts ---
\bibitem{charaka_samhita}
Charaka. (circa 100 AD). \emph{Charaka Samhita}. Ancient Ayurvedic Compendium.

\bibitem{sushruta_samhita}
Sushruta. (circa 200 AD). \emph{Sushruta Samhita}. Ancient Ayurvedic Compendium.

\bibitem{ashtanga_hridaya}
Vagbhata. (circa 600 AD). \emph{Ashtanga Hridaya}. Classical Ayurveda text.

\bibitem{kashyapa_samhita}
Kashyapa. (circa 600 AD). \emph{Kashyapa Samhita}. Pediatrics-focused Ayurvedic Compendium.

% --- Indic Corpora & Resources ---
\bibitem{archiveorg}
Internet Archive. (2025). \emph{Archive.org Digital Library}. \url{https://archive.org}

\bibitem{egangotri}
eGangotri Foundation. (2025). \emph{eGangotri Digital Library}. \url{https://egangotri.in}

\bibitem{ndli}
NDLI. (2025). \emph{National Digital Library of India}. \url{https://ndl.iitkgp.ac.in}

\bibitem{ai4bharat_corpora}
Kakwani, D., et al. (2020). IndicCorp and XL-Sum: Benchmarking Multilingual Corpora for Indian Languages. 
\emph{Findings of ACL 2020}.

\bibitem{indicnlp}
Kunchukuttan, A. (2020). The IndicNLP Library: Natural Language Processing for Indian Languages. 
\emph{EMNLP Workshop}.

\bibitem{mc4}
Xue, L., et al. (2021). mC4: A Multilingual C4 Corpus. 
\emph{TACL}.

% --- Large-scale LLM Datasets ---
\bibitem{pile}
Gao, L., et al. (2021). The Pile: An 800GB Dataset of Diverse Text for Language Modeling. 
\emph{NeurIPS Datasets and Benchmarks}.

\bibitem{roots}
Laurençon, H., et al. (2022). The ROOTS Corpus: A 1.6TB Multilingual Dataset for Training Large Language Models. 
\emph{ACL}.

\bibitem{dolma}
Soldaini, L., et al. (2023). Dolma: A Large-Scale, Long-Context, English Dataset. 
\emph{arXiv preprint arXiv:2306.07328}.

\bibitem{refinedweb}
Penedo, G., et al. (2023). RefinedWeb: A Large-Scale Dataset for Web-Scale Language Models. 
\emph{NeurIPS}.

\bibitem{c4}
Raffel, C., et al. (2020). Exploring the Limits of Transfer Learning with a Unified Text-to-Text Transformer. 
\emph{JMLR}.

\bibitem{bookscorpus}
Zhu, Y., et al. (2015). BookCorpus: A Large Collection of Free Books. 
\emph{ICCV Workshop}.

\bibitem{openwebtext}
Gokaslan, A., \& Cohen, V. (2019). OpenWebText Corpus. 
\emph{Open Source Release}.

% --- OCR ---
\bibitem{surya_ocr}
Mishra, A., \& Sharma, R. (2023). Surya: Multilingual OCR for Indic Scripts. 
\emph{Proceedings of the Indic NLP Symposium}.

% --- Ethics & Governance ---
\bibitem{gebru_datasheets}
Gebru, T., et al. (2018). Datasheets for Datasets. 
\emph{FAT* Conference}.

\bibitem{model_cards}
Mitchell, M., et al. (2019). Model Cards for Model Reporting. 
\emph{FAT* Conference}.

\bibitem{dataset_ethics}
Gebru, T., et al. (2021). Ethical Considerations in Dataset Collection. 
\emph{FAccT Conference}.

\bibitem{cc0}
Creative Commons. (2010). Creative Commons Zero License (CC0). 
\url{https://creativecommons.org/publicdomain/zero/1.0/}

\bibitem{cc_by}
Creative Commons. (2010). Creative Commons Attribution License (CC-BY). 
\url{https://creativecommons.org/licenses/by/4.0/}

% --- Other Foundation Models ---
\bibitem{qwen3}
Alibaba DAMO Academy. (2024). Qwen3-235B: Large Language Model. 
\emph{Technical Report}.

\bibitem{gemma}
Google DeepMind. (2024). Gemma: Open Language Models by Google DeepMind. 
\emph{Technical Report}.

\end{thebibliography}

\end{document}